\documentclass[11pt]{article}
\usepackage{times}
\usepackage{CJKutf8}
\usepackage{color}
\usepackage{enumerate}
\usepackage{algorithm}
\usepackage{algorithmic}
\usepackage{multicol}
\usepackage{amssymb, amsmath}
\usepackage{stmaryrd}
\usepackage{graphicx}
\usepackage{setspace}
\usepackage{mathrsfs}
\usepackage[bookmarks=true]{hyperref}

\newtheorem{definition}{Definition}
\newtheorem{proposition}{Proposition}

\newcommand{\defeq}{\stackrel{\mathrm{def}}{=}}

\newcommand{\bfalpha}{\boldsymbol{\alpha}}

\newcommand{\bfgamma}{\boldsymbol{\gamma}}

\newcommand{\bfnu}{\boldsymbol{\nu}}
\newcommand{\bfomega}{\boldsymbol{\omega}}
\newcommand{\bfsigma}{\boldsymbol{\sigma}}
\newcommand{\bftau}{\boldsymbol{\tau}}

\newcommand{\bff}{\mathbf{f}}

\newcommand{\bfh}{\mathbf{h}}

\newcommand{\bfn}{\mathbf{n}}

\newcommand{\bfq}{\mathbf{q}}

\newcommand{\bfu}{\mathbf{u}}

\newcommand{\bfD}{\mathbf{D}}

\newcommand{\bfI}{\mathbf{I}}
\newcommand{\bfJ}{\mathbf{J}}
\newcommand{\bfJw}{\mathbf{J}_\mathrm{wr}}

\newcommand{\bfM}{\mathbf{M}}

\newcommand{\bfP}{\mathbf{P}}

\newcommand{\bfS}{\mathbf{S}}

\newcommand{\bfW}{\mathbf{W}}

\newcommand{\calS}{{\cal S}}

\def\ie{\emph{i.e.},~}
\def\eg{\emph{e.g.},~}
\def\q{\bfq}
\def\qd{\dot{\q}}
\def\qdd{\ddot{\q}}
\def\taux{\tau^x}
\def\tauy{\tau^y}
\def\tauz{\tau^z}

\def\Jrot{\bfJ_{\mathrm{rot}}}
\def\tausafe{\tau_\textrm{safe}}
\def\taumin{\tau_\textrm{min}}
\def\taumax{\tau_\textrm{max}}
\def\sgn{\textrm{sgn}}

\renewcommand\d[1]{{\rm d}{#1}}

\usepackage{authblk}

\title{\Large Stability of Surface Contacts for Humanoid Robots:\\ 
  Closed-Form Formulae of the Contact Wrench Cone\\
  for Rectangular Support Areas}
  
\author[1]{St\'ephane Caron}%
\author[2]{Quang-Cuong Pham}%
\author[1]{Yoshihiko Nakamura}

\affil[1]{Department of Mechano-Informatics, The University of Tokyo,
  Japan}%
\affil[2]{School of Mechanical and Aerospace Engineering, Nanyang
  Technological University, Singapore}

\begin{document}

\maketitle

\begin{abstract}
    Humanoid robots locomote by making and breaking contacts with their
    environment. A crucial problem is therefore to find precise criteria for
    a given contact to remain stable or to break. For rigid surface contacts,
    the most general criterion is the Contact Wrench Condition (CWC). To check
    whether a motion satisfies the CWC, existing approaches take into account
    a large number of individual contact forces (for instance, one at each
    vertex of the support polygon), which is computationally costly and
    prevents the use of efficient inverse-dynamics methods. Here we argue that
    the CWC can be explicitly computed without reference to individual contact
    forces, and give closed-form formulae in the case of rectangular surfaces
    -- which is of practical importance. It turns out that these formulae
    simply and naturally express three conditions: (i) Coulomb friction on the
    resultant force, (ii) ZMP inside the support area, and (iii) bounds on the
    yaw torque. Conditions (i) and (ii) are already known, but condition (iii)
    is, to the best of our knowledge, novel. It is also of particular interest
    for biped locomotion, where undesired foot yaw rotations are a known issue.
    We also show that our formulae yield simpler and faster computations than
    existing approaches for humanoid motions in single support, and demonstrate
    their consistency in the OpenHRP simulator.
\end{abstract}

\section{Introduction}

From the viewpoint of the robot, establishing contact with the environment
amounts to constraining a certain number of Degrees Of Freedom (DOF) of an
end-effector link. For instance, a sliding planar contact imposes three
\emph{equality} constraints (two on the orientation of the link and one for the
link-to-surface distance) while a fixed contact constraints all six DOFs of the
end-effector link. A stability criterion can be seen as a set of
\emph{inequality} constraints describing the conditions under which these
equalities are preserved.

The venerable Zero-Moment Point (ZMP) \cite{vukobratovic2004zero} criterion may
be the best-known example of stability criterion for rigid surface contact.  It
is known \cite{hirukawa2006universal} to be necessary but \emph{not sufficient}
for contact stability: in particular, it says nothing about possible sliding in
the horizontal plane and rotation around the vertical axis (yaw). Yet, humanoid
robots are often subject to significant yaw moments during single support
phases, resulting in undesired foot rotations. Attesting the importance of this
problem, recent works have been using upper-body motions to compensate for
these yaws while continuing to use ZMP \cite{ugurlu2012yaw, cisneros2014yaw}.

A more principled (and general) way to address this problem is to consider
\emph{individual contact forces} distributed on the contact surface, as can be
found \eg in bipedal balance control \cite{ott2011posture} and motion planning
\cite{saab2013dynamic, hauser2014fast}. This approach yields a stronger
stability criterion than ZMP, and accounts for both the sliding and yaw
rotation. It is however hampered by \emph{redundancy}: the vector of contact
forces has many more components (three times the number of contact points) than
the degree of the contact constraint (six). This redundancy makes the
resolution of the equations of motion fundamentally harder, as illustrated by
the fact that state-of-the-art Inverse Dynamics based on QR-decomposition
\cite{righetti2013optimal} only apply to non-redundant variables\,\footnote{
    When this is not the case, the authors advocate the use of Singular Value
    Decomposition to compute new independent variables; however, it is unclear
    how to compute the inequality constraints applying to these new variables.
}. In Time-Optimal Path Parameterization (TOPP), redundancy prompted the use of
further contact approximations \cite{caron2014kinodynamic} or expensive
polytope projection algorithms \cite{hauser2014fast}. In the present paper, we
argue that such workarounds are no longer necessary if one uses the correct
contact representation (for instance, both \cite{caron2014kinodynamic} and
\cite{hauser2014fast} boil down to a single matrix inversion for a biped in
single-support, as we will see in Section \ref{sec:experiment}).

The key insight here is that the condition that individual contact forces lie
in their respective friction cones can be replaced by the condition that the
\emph{contact wrench} belongs to a certain \emph{wrench
cone}~\cite{balkcom2002computing}. The contact wrench naturally solves the redundancy
issue, as its dimension is minimal (six). It was advocated as a generalization
of ZMP in \cite{hirukawa2006universal}, along with a stability theorem, and
applied to walking pattern generation on rough terrains
\cite{hirukawa2007pattern, zheng2010walking}. However, this theorem makes the
same ``sufficient friction'' assumption as ZMP, which means the resulting
criterion does not account for sliding and yaw rotations. Besides, the contact
wrench is computed from individual contact forces, which we argue yields
unnecessarily extensive calculations: following~\cite{balkcom2002computing}, the
wrench cone can be computed explicitly from the sole geometry of the contact
surface.

In this paper, we derive the closed-form formulae of the wrench cone in the
case of rectangular support areas, which is of practical importance since most
humanoid robot feet can be adequately approximated by rectangles. This result
helps simplify dynamics computations, as we will see for humanoid motions. It
also provides an analytical description of \emph{``yaw friction''}, from which
we derive a simple and principled control law to prevent undesirable yaw
rotations.

The rest of the paper is organized as follows. In Section \ref{sec:background},
we recall the definitions related to contact stability. In Section
\ref{sec:surface}, we discuss the physics of surface contact and give
a theoretical justification for the practice of considering individual contact
forces at the vertices of the contact polygon. Then, in Section \ref{sec:cwc},
we derive a closed-form expression of the wrench cone in the case of
rectangular contact areas. We apply the resulting solution in a humanoid
experiment in Section \ref{sec:experiment} before concluding in Section
\ref{sec:conclusion}.

\section{Background}
\label{sec:background}

\subsection{Contact Forces and Contact Wrench}

Consider a robot with $n$ degrees of freedom making $N$ point contacts with the
environment, at points $C_1,\dots,C_N$ in the laboratory reference frame (RF).
The equations of motion of the robot are:
\begin{equation}
    \label{eq:motion}
    \bfM(\q) \qdd + \bfh(\q, \qd) = \bfS^\top \bftau_a + \sum_{i=1}^N
    \bfJ(C_i)^\top \bff_i,
\end{equation}
where $\bfq, \qd, \qdd$ are the $n$-dimensional vectors of DOF values,
velocities and accelerations, $\bfM$ is the $n \times n$ inertia matrix,
$\bfh(\q, \qd)$ the $n$-dimensional vector of gravity and Coriolis forces. In
case the robot has $n_a$ actuated joints, $\bftau_a$ is the $n_a$-dimensional
vector of torques at the actuated joints and $\bfS$ is a $n_a \times n$ joint
selection matrix. Finally, for each $i\in[1, N]$, $\bff_i$ is a 3-dimensional
vector of contact force and $\bfJ(C_i)$ is the $3 \times n$ \emph{translation}
Jacobian calculated at point $C_i$.

We assume that both the environment and the contacting link are rigid bodies.
Thus, interactions between them can be represented by a single \emph{contact
wrench} $\bfW~=~(\bff, \bftau)$, which we can compute from contact forces as:
\begin{eqnarray}
    \bff & \defeq & \sum_i \bff_i, \label{Wf-f} \\
    \bftau & \defeq & \sum_i \overrightarrow{OC_i}\wedge \bff_i, \label{Wf-tau}
\end{eqnarray}
where $O$ is the origin of the link RF. The contact jacobian $\bfJw$ is the
$6 \times n$ matrix obtained by stacking vertically $\bfJ(O)$, the translation
Jacobian computed at $O$, and $\Jrot$, the \emph{rotation} Jacobian of the
link, both taken with respect to the absolute RF. With these definitions, we
have the following property:
\begin{equation}
    \label{equaJ}
    \bfJw^\top \bfW = \sum_{i=1}^N \bfJ(C_i)^\top \bff_i.
\end{equation}
(See Appendix \ref{proof:equivwrench} for a proof.) Then, the equations of
motion can be rewritten as:
\begin{equation}
    \label{eq:motionwrench}
    \bfM(\q) \qdd + \bfh(\q, \qd) = \bfS^\top \bftau_a + \bfJw^\top \bfW.
\end{equation}

\subsection{Contact Stability}

\begin{figure}[thp]
    \centering
    \includegraphics[width=4cm]{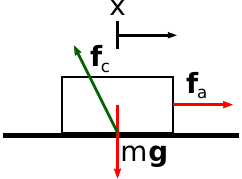}
    \caption{Block on a horizontal surface with one DOF.}
    \label{fig:onedof}
\end{figure}

Assume now that the robot is in a given \emph{state} $(\bfq,\qd)$. The
accelerations $\qdd$ and generalized forces exerted on the robot (actuated
torques or contact forces) are bound by a \emph{complementarity} conditions
\cite{pang2000stability}. For the sake of the explanation, let us consider
first the simple case of a single translation coordinate $x$, as depicted in
Figure~\ref{fig:onedof}.

\noindent Under Coulomb's friction model, either of the following situations
occurs\,:
\begin{itemize}

    \item \textbf{Fixed contact:} $\ddot x=0$ and the contact force obeys
        $|f_c^t| \leq \mu f_c^n$, where $f_c^t$ and $f_c^n$ are respectively
        the horizontal and vertical components of $\bff_c$ and $\mu$ is the
        static friction coefficient;

    \item \textbf{Sliding contact:} $\ddot x>0$ and the contact force
        obeys $f_c^t = -\mu_k f_c^n$, where $\mu_k$ is the kinetic friction
        coefficient.

\end{itemize}
The acceleration $\ddot x$ and contact force $\bff_c$ are in
a \emph{complementary} relationship: when one is equality-constrained, the
other is inequality-constrained. Similarly, for the general case of a 6-DOF
end-effector in contact, the translational and rotational accelerations of the
link are in a complementary relationship with some generalized contact forces
$\bfgamma$ (here, $\bfgamma = \bfW$ or $(\bff_1, \ldots, \bff_N)$).  The
\emph{contact mode} describes, for each variable in a complementary
relationship, whether it is equality- or inequality-constrained. In contact
stability, we are interested in the \emph{fixed contact mode} where the
position and orientation of the end-effector are equality-constrained to
a reference value. 

\begin{definition}[Weak Contact Stability]
    A contact is weakly stable when there exists a solution $(\qdd, \bftau_a,
    \bfgamma)$ of the equations of motion satisfying the fixed contact mode for
    all contacting links. 
\end{definition}

That is to say, 
\begin{itemize}
    \item for each contact $(i)$, the relative velocity and acceleration at contact are zero:
        $\bfJw^{(i)} \qd = 0$ and $\bfJw^{(i)} \qdd = - \dot{\bfJ}_\textrm{wr}^{(i)} \qd$,
    \item actuated torques $\bftau_a$ are within torque limits,
    \item complementary forces $\bfgamma$ satisfy their inequality constraints (friction
        cones or wrench cone).
\end{itemize}
This formulation has been widely used in the literature. In approaches based on
inverse dynamics, the conditions on $(\qd, \qdd)$ are first enforced
kinematically, then torques and complementary forces are solved
\cite{hauser2014fast, righetti2013optimal}.

The ``weakness'' in the definition above refers to the notions of strong and
weak stability, as stated by \cite{pang2000stability}. Strong stability happens
when \emph{all} solutions to the equations of motion satisfy the fixed contact
mode. Note that choosing between contact forces and the contact wrench changes
the equations of motion (Equations \eqref{eq:motion} and
\eqref{eq:motionwrench}, respectively), but the underlying contact stability is
the same by Equation \eqref{equaJ}. In the rest of the paper, we will always
refer to contact stability in the weak sense.

\section{Surface Contact}
\label{sec:surface}

Suppose we take contact forces $\bff_1, \ldots, \bff_N$ as complementary
variables to the position and orientation of the contacting link. Let $f_i^{n}$
and $\bff_i^{t}$ denote the normal and the tangential components of the contact
force $\bff_i$. Coulomb friction provides the complementary inequalities:
\vspace{.1em}
\begin{itemize}

    \item \textbf{Unilaterality:} 
        \vspace{-1.1em}
        \begin{equation}
            \label{comp-force-1} f_i^{n} > 0,
        \end{equation}

    \item \textbf{Non-slippage:} 
        \vspace{-1.2em}
        \begin{equation}
            \label{comp-force-2} \|\bff_i^{t}\| \leq \mu f_i^{n},
        \end{equation}

\end{itemize}
\vspace{.1em}
where $\mu$ is the static coefficient of friction. 

\begin{figure}[thp]
  \centering
  \includegraphics[width=0.75\columnwidth]{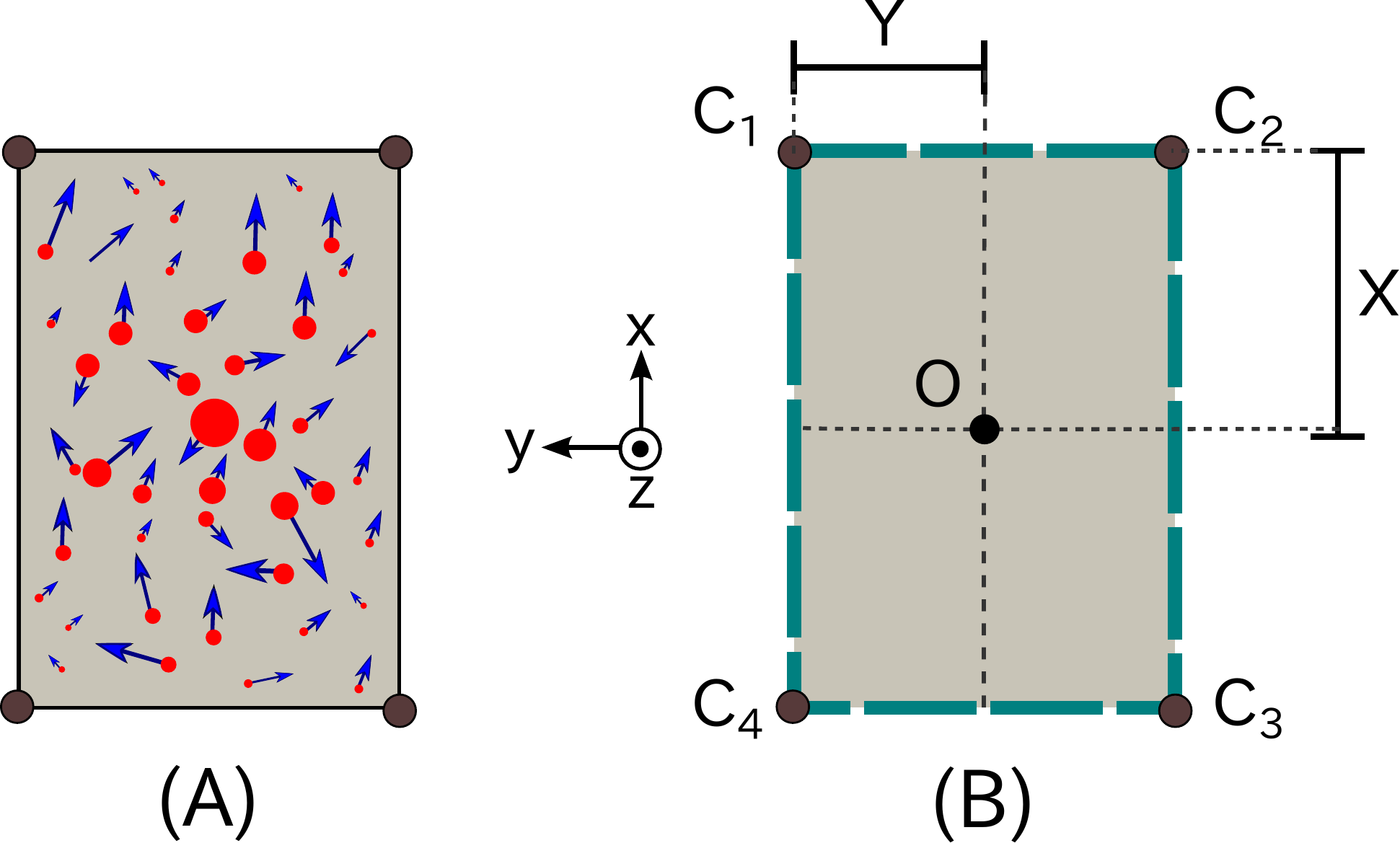}
  \caption{
      Contact in the surface plane. (A) Example of stress/pressure fields. Red
      discs indicate the magnitude of pressure by their size. Blue arrows show
      tangential stress. (B) Notations used in Section \ref{sec:cwc}.
  }
  \label{fig:foot}
\end{figure}

However, when the contact is done through a surface and not through a set of
points, the reality of contact is that of continuum mechanics. In this case,
the action of the environment at the contact surface $\calS$ is described by
two quantities: a scalar field $p(x, y)$ corresponding to normal
\emph{pressure}, and a two-dimensional vector field $\bfsigma(x, y)$ for
tangential mechanical \emph{stress}. Figure \ref{fig:foot}-(A) illustrates
these two fields for a rectangular contact area. For convenience, we also will
denote by $\bfnu \defeq \bfsigma(x,y) + p(x,y) \bfn$, where $\bfn$ is the unit
vector normal to the contact surface. The equations of motion become
\begin{eqnarray}
    \label{eq:motioncont}
    \bfM \qdd + \bfh = \bfS^\top \bftau_a + 
    \int_\calS
    \bfJ(C_{xy})^\top \bfnu(x, y) \d{x}\d{y}
\end{eqnarray}
where $\bfJ(C_{xy})$ is the $3 \times n$ translation Jacobian calculated at the
point of coordinate $C_{xy}$ on the surface (taken in the laboratory RF). The
wrench resulting from $\bfnu(x, y)$ is
\begin{eqnarray}
    \bff & \defeq & \int_\calS \bfnu(x,y) \d{x}\d{y}, 
    \label{Wsurf-f} \\
    \bftau & \defeq & \int_\calS \overrightarrow{OC_{xy}}\wedge \bfnu(x,y) \d{x}\d{y},
    \label{Wsurf-tau}
\end{eqnarray}
where $O$ is the origin of the link RF. Under Coulomb friction, the inequality
constraints for $\bfnu(x, y)$ are:
\begin{itemize}

    \item \textbf{Unilaterality:} 
        \vspace{-1.1em}
        \begin{equation}
            \label{comp-surf-1} p(x, y) > 0,
        \end{equation}

    \item \textbf{Non-slippage:}
        \vspace{-1.8em}
        \begin{equation}
            \label{comp-surf-2} \| \bfsigma(x, y) \| \leq \mu\,p(x, y).
        \end{equation}

\end{itemize}
\vspace{.1em}
Note that taking a constant $\mu$ in Equation \eqref{comp-surf-2} is an
approximation: in reality, the friction coefficient $\mu(x, y)$ varies with the
position on the surface.

In the present literature, surface contact is often modeled using sets of
contact points: sometimes more than required by positional constraints (\eg
nearly 30 per contacting link in \cite{hauser2014fast}) or one at each vertex
of the convex hull (\eg in \cite{ott2011posture}). The proposition below gives
a theoretical justification for the latter practice.

\begin{proposition}
    \label{prop:componeway}
    Assume that the contact surface $\calS$ is a convex polygon with vertices
    $C_1, \ldots, C_N$. If there exists $\bfnu(x, y)_{(x, y) \in \calS}$
    satisfying complementary inequality constraints
    \eqref{comp-surf-1}-\eqref{comp-surf-2}, then there will exist contact
    forces applied at $C_1, \ldots, C_N$ and summing up to the same contact
    wrench that satisfy complementary inequality constraints
    \eqref{comp-force-1}-\eqref{comp-force-2}.
\end{proposition}

\emph{Proof:}\,consider pressure and stress fields summing up to $\bfW$. By
convexity, one can find strictly positive functions $\alpha_1(x, y), \ldots,
\alpha_k(x, y)$ such that $\sum_i\alpha_i(x, y) = 1$ and each point $C_{xy}
\in \calS$ can be written $C_{xy} = \sum_i \alpha_i(x, y) C_i$. Then, define
for each vertex $C_i$ a force
\begin{equation*}
    \bff_i := \int_{\calS} \alpha_i(x, y) \bfnu(x,y) \d{x} \d{y},
\end{equation*}
By positivity of the $\alpha_i$'s, it is straightforward to check that all
$f_i^n>0$ and $\|\bff_i^t\| \leq \mu f_i^n$. In addition, this expression of
$\bff_i$ ensures that the resulting wrenches are equal, \ie \eqref{Wf-f} $=$
\eqref{Wsurf-f} and \eqref{Wf-tau} $=$ \eqref{Wsurf-tau}. $\Box$

We furthermore argue that, when the friction coefficient $\mu$ is assumed to be
constant as in Equation \eqref{comp-surf-2}, the two conditions are equivalent.
The complete proof of this equivalence, which involves reconstructing pressure
and stress fields given local and boundary conditions, is however out of the
scope of this paper. The bottom line of this argument is that using contact
forces at vertices of the convex hull completely describes the dynamics of
surface contact. The wrench cone that we derive in the next section will share
the same characteristic.

\section{Wrench Cone for Rectangular Surfaces}
\label{sec:cwc}

Consider a rectangular area $(C_1C_2C_3C_4)$ as depicted in
Figure~\ref{fig:foot} (B). We calculate the contact wrench $\bfW$ at the center
$O$ in the link's RF.\footnote{multiply by the link's rotation matrix for
a wrench in the absolute RF} Let us denote by $(f_i^x, f_i^y ,f_i^z)$ the three
components of the contact force at point $C^i$. Under the common linear
approximation of friction cones, Coulomb inequalities become
\begin{eqnarray}
    |f_i^x|, |f_i^y| & \leq & \mu f_i^z \label{fixy} \\
    f_i^z & > & 0 \label{fiz}
\end{eqnarray}
The wrench cone is then given by the following proposition.

\begin{proposition}
    There exists a solution $(\bff_1, \ldots, \bff_4)$ satisfying inequalities
    \eqref{fixy}-\eqref{fiz} if and only if there exists a wrench $\bfW = (f^x,
    f^y, f^z, \taux, \tauy, \tauz)$ such that:
    \begin{eqnarray}
        |f^x| & \leq & \mu f^z                    \label{W1} \\
        |f^y| & \leq & \mu f^z                    \label{W2} \\
        f^z & > & 0                               \label{W3} \\
        |\taux| & \leq & Y f^z                    \label{W4} \\
        |\tauy| & \leq & X f^z                    \label{W5} \\
        \taumin \leq & \tauz & \leq \ \taumax     \label{W6}
    \end{eqnarray}
    where \small 
    \begin{eqnarray*}
        \taumin & \defeq & -\mu (X+Y) f^z + |Y f^x - \mu \taux| + |X f^y - \mu \tauy|, \\
        \taumax & \defeq & +\mu (X+Y) f^z - |Y f^x + \mu \taux| - |X f^y + \mu \tauy|.
    \end{eqnarray*}
\end{proposition}

\vspace{.5em}

The proof of this proposition is given in Appendix \ref{bigproof}. The validity
of this expression was also tested empirically with a script available at
\cite{webpage}.

Let us now detail each line of the wrench cone. The first two inequalities
\eqref{W1}-\eqref{W2} correspond to the usual Coulomb friction.  Inequalities
\eqref{W3}, \eqref{W4} and \eqref{W5} are equivalent to the ZMP condition. The
last inequality \eqref{W6} provides a bound on the admissible yaw torque that
was implicitely encoded in the contact-force model. Note how this relation is
more complex than a mere ``no rotation occurs while $\tauz$ is small enough'',
as it is coupled with all other components of the contact wrench. Notably, the
``safest'' value is not zero but:
\begin{eqnarray*}
    \tausafe & \defeq & (\taumin + \taumax) / 2 \\
            & = & \sgn(-f^x \taux) \min(Y |f^x|, \mu |\taux|) \\
            & + & \sgn(-f^y \tauy) \min(X |f^y|, \mu |\tauy|),
\end{eqnarray*}
where $\sgn$ is the sign function. From \eqref{W6}, $\tauz$ may deviate from
$\tausafe$ by at most
\[
    \mu (X+Y) f^z - \max(Y |f^x|, \mu |\taux|) - \max(X |f^y|, \mu |\tauy|).
\]
We see that higher tangential forces or roll-pitch torques reduce the range of
admissible yaw torques. In particular, when these other constraints are
saturated (for instance when the ZMP reaches a corner of the support polygon),
$\tausafe$ is the \emph{only solution} that prevents the contact from breaking.
Therefore, $\tauz = \tausafe$ appears as a sensible control law to prevent
undesired yaw rotations.

    \begin{figure*}
        \centering
        \includegraphics[width=.99\textwidth]{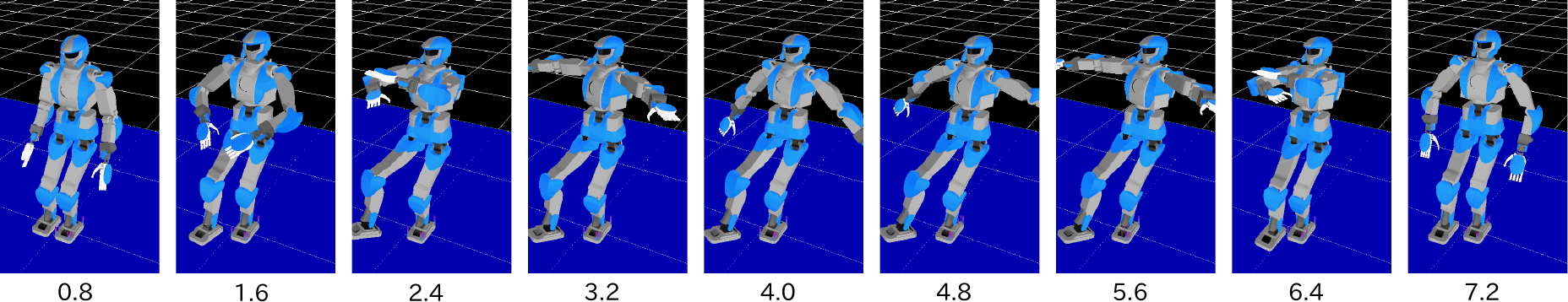}
        \caption{
            Snapshots of the retimed motion. The motion lasts 7.285 s. Time
            stamps are shown below each frame. This motion stresses all
            components of the wrench cone. The first part stresses the pitch by
            moving the COM forward. The second segment extends the arms, then
            keeps the left arm still while moving the right arm, thus stressing
            the roll. At the same time, the chest pitch is actuated back and
            forth, which stresses the yaw. Finally, the waist performs an
            elliptic motion (back and forth, up and down) throughout the whole
            motion, thus stressing the translation of the contact foot.
        }
        \label{fig:retimed}
    \end{figure*}

    \begin{figure*}
        \centering
        \includegraphics[width=.99\textwidth]{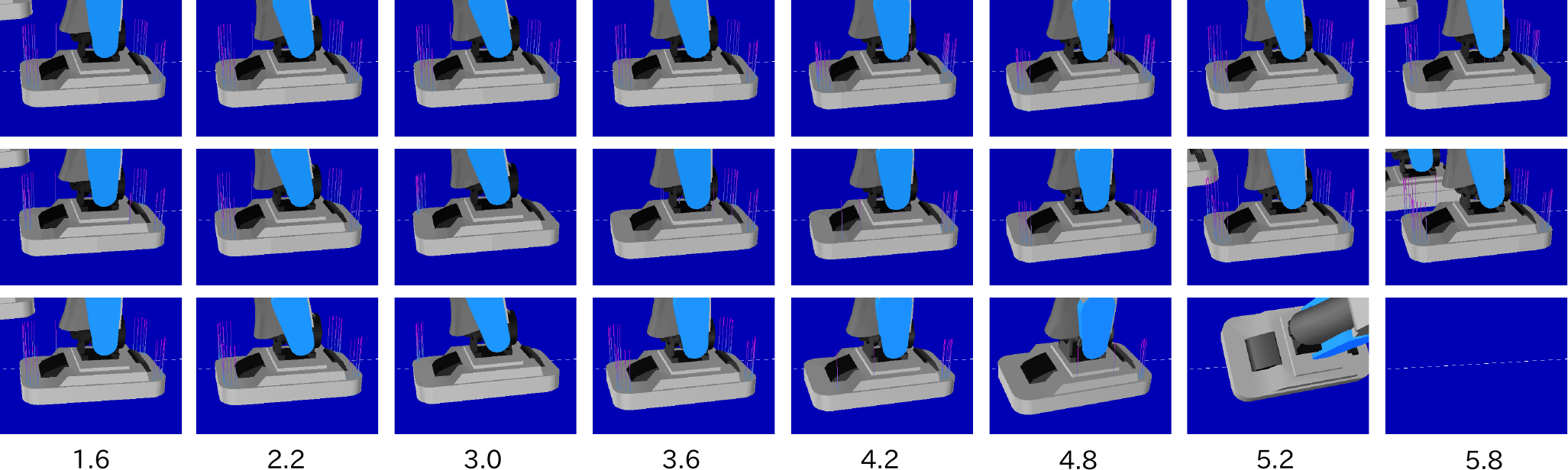}
        \caption{
            Zoom on the contact forces computed by OpenHRP at the left foot.
            The first row corresponds to the retimed motion (7.29 s), the
            second to the 10\% uniform acceleration of the retimed motion (6.55
            s) and the third to the 15\% uniform acceleration (6.19 s). Time
            stamps are shown below each column. The retimed motion maintains
            surface contact with contact forces distributed all around the
            contact area, except between 4.2 and 4.8 s where they are mostly on
            the left half of the foot. The 10\%-accelerated version goes to
            point contact (3.0 s) and line edge contact (3.6 to 4.8 s) but does
            not fall. At 15\% acceleration, the CWC violation overcomes the
            stabilizer's ability and the robot falls.
        }
        \label{fig:footzoom}
    \end{figure*}

\section{Experiment}
\label{sec:experiment}

We test the validity of the Contact Wrench Condition (CWC) in the integrated
simulator OpenHRP with a model of the HRP4 robot. Note that OpenHRP has its own
contact model where forces are distributed along the edges (not corners) of the
contact surface.

We implemented the CWC condition within the Time-Optimal Path Parameterization
framework (TOPP), a well-known projection of system dynamics along
a pre-defined path that has been used for motion planning of humanoid robots
\cite{hauser2014fast, pham2013kinodynamic, caron2014kinodynamic}. We considered
the case of a single contact at the left foot and designed a motion that would
challenge all six contact DOFs. In single contact, the contact wrench is fully
determined by the unactuated rows of the equation of motion, \ie
\begin{equation*}
    \bfW = (\bfP \bfJw)^{-1} \bfP \left[ \bfM(\q) \qdd + \bfh(\q, \qd) \right]
\end{equation*}
where $\bfP = \bfI - \bfS$ is the unactuated line selection matrix. Note how
computing the constraint projector is straightforward when using the contact
wrench: it is not the case when contact forces are used, as we observed in
previous research \cite{caron2014kinodynamic} (where we resolved the force
redundancy with a force-binding model). 


For this experiment, we designed by hand a set of eleven key postures. The
geometric path was obtained by interpolating Bezier curves between these
postures and applying a kinematic filter to fix the position and orientation
of the support foot on the ground. To get a feasible trajectory from this path
(\ie to compute the timing information) we used the open-source TOPP solver
\cite{pham2014general}. The solver takes as input the path and a vector
representation of the system dynamics along it, which is easy to compute once
one knows the projectors mentioned above (see \cite{caron2014kinodynamic} for
details). Figure~\ref{fig:retimed} shows a timelapse of the final retimed
motion. Videos are also available online at \cite{webpage}.

An interesting thing to note about TOPP is that, because of time optimality,
the retimed motion always saturates at least one of the contact constraints. In
an ideal setting with perfect system dynamics, the obtained trajectory should
therefore be at the limit of contact: it would execute correctly, but the
contact would break as soon as one tries to perform the motion faster at any
time instant. We observed this phenomenon in the experiment, as depicted in
Figure~\ref{fig:footzoom}. 

There are, however, a few points to discuss. In order to ensure proper tracking
of the computed trajectory, we used HRP4's stabilizer module, which
occasionally changes the actual joint angles. In turn, this behavior violates
the assumption made by TOPP that the robot follows a precise geometric path,
and the constraints may get under- or over-saturated. To alleviate for this
issue, we added safety margins in TOPP's conditions: we scaled the contact area
by 45\% and set a smaller friction coefficient $\mu = 0.4$ (versus $\mu = 0.8$
in OpenHRP). Consequently, it was possible in practice to accelerate the
retimed motion by about 5\% (\ie reducing the total duration by 5\% with
uniform timescaling) and obtain a successful execution.  However, we still
observed the expected phenomenon with relatively small changes: for a 10\%
acceleration (by uniform timescaling), the humanoid started to lose surface
contact, but the stabilizer was still able to recover; for a 15\% acceleration,
the violation of the contact constraint was too large and the robot fell. See
Figure~\ref{fig:footzoom}.

\section{Conclusion}
\label{sec:conclusion}

In this paper, we calculated the closed-form expression of the wrench cone for
rectangular contact surfaces. This formula has several implications. First, it
is very simple, making computations much easier than any previous formulation
based on contact forces. Second, it describes concisely the phenomenon of
\emph{yaw friction} by a double-inequality that is, to the best of our
knowledge, novel. From these bounds, we derived a simple control law to avoid
undesired foot rotations, a recurring problem for bipeds in single contact.
Finally, we showed how our criterion can give simpler and faster computations
than contact forces for humanoid motions in single support, and demonstrated it
with dynamic motions simulated in OpenHRP.

{\scriptsize
\bibliographystyle{IEEEtran}
\bibliography{refs}}

\begin{thebibliography}{10}
\providecommand{\url}[1]{#1}
\csname url@samestyle\endcsname
\providecommand{\newblock}{\relax}
\providecommand{\bibinfo}[2]{#2}
\providecommand{\BIBentrySTDinterwordspacing}{\spaceskip=0pt\relax}
\providecommand{\BIBentryALTinterwordstretchfactor}{4}
\providecommand{\BIBentryALTinterwordspacing}{\spaceskip=\fontdimen2\font plus
\BIBentryALTinterwordstretchfactor\fontdimen3\font minus
  \fontdimen4\font\relax}
\providecommand{\BIBforeignlanguage}[2]{{%
\expandafter\ifx\csname l@#1\endcsname\relax
\typeout{** WARNING: IEEEtran.bst: No hyphenation pattern has been}%
\typeout{** loaded for the language `#1'. Using the pattern for}%
\typeout{** the default language instead.}%
\else
\language=\csname l@#1\endcsname
\fi
#2}}
\providecommand{\BIBdecl}{\relax}
\BIBdecl

\bibitem{vukobratovic2004zero}
M.~Vukobratovi{\'c} and B.~Borovac, ``Zero-moment point—thirty five years of
  its life,'' \emph{International Journal of Humanoid Robotics}, vol.~1,
  no.~01, pp. 157--173, 2004.

\bibitem{hirukawa2006universal}
H.~Hirukawa, S.~Hattori, K.~Harada, S.~Kajita, K.~Kaneko, F.~Kanehiro,
  K.~Fujiwara, and M.~Morisawa, ``A universal stability criterion of the foot
  contact of legged robots-adios zmp,'' in \emph{Robotics and Automation, 2006.
  ICRA 2006. Proceedings 2006 IEEE International Conference on}.\hskip 1em plus
  0.5em minus 0.4em\relax IEEE, 2006, pp. 1976--1983.

\bibitem{ugurlu2012yaw}
B.~Ugurlu, J.~A. Saglia, N.~G. Tsagarakis, and D.~G. Caldwell, ``Yaw moment
  compensation for bipedal robots via intrinsic angular momentum constraint,''
  \emph{International Journal of Humanoid Robotics}, vol.~9, no.~04, 2012.

\bibitem{cisneros2014yaw}
R.~Cisneros, K.~Yokoi, and E.~Yoshida, ``Yaw moment compensation by using full
  body motion,'' in \emph{Mechatronics and Automation (ICMA), 2014 IEEE
  International Conference on}.\hskip 1em plus 0.5em minus 0.4em\relax IEEE,
  2014, pp. 119--125.

\bibitem{ott2011posture}
C.~Ott, M.~A. Roa, and G.~Hirzinger, ``Posture and balance control for biped
  robots based on contact force optimization,'' in \emph{Humanoid Robots
  (Humanoids), 2011 11th IEEE-RAS International Conference on}.\hskip 1em plus
  0.5em minus 0.4em\relax IEEE, 2011, pp. 26--33.

\bibitem{saab2013dynamic}
L.~Saab, O.~E. Ramos, F.~Keith, N.~Mansard, P.~Soueres, and J.~Fourquet,
  ``Dynamic whole-body motion generation under rigid contacts and other
  unilateral constraints,'' \emph{Robotics, IEEE Transactions on}, vol.~29,
  no.~2, pp. 346--362, 2013.

\bibitem{hauser2014fast}
K.~Hauser, ``Fast interpolation and time-optimization with contact,'' \emph{The
  International Journal of Robotics Research}, vol.~33, no.~9, pp. 1231--1250,
  2014.

\bibitem{righetti2013optimal}
L.~Righetti, J.~Buchli, M.~Mistry, M.~Kalakrishnan, and S.~Schaal, ``Optimal
  distribution of contact forces with inverse-dynamics control,'' \emph{The
  International Journal of Robotics Research}, vol.~32, no.~3, pp. 280--298,
  2013.

\bibitem{caron2014kinodynamic}
S.~Caron, Y.~Nakamura, and Q.-C. Pham, ``Kinodynamic motion retiming for
  humanoid robots,'' in \emph{Proceedings of the 32nd Annual Conference of the
  Robotics Society of Japan}, 2014.

\bibitem{balkcom2002computing}
D.~J. Balkcom and J.~C. Trinkle, ``Computing wrench cones for planar rigid body
  contact tasks,'' \emph{The International Journal of Robotics Research},
  vol.~21, no.~12, pp. 1053--1066, 2002.

\bibitem{hirukawa2007pattern}
H.~Hirukawa, S.~Hattori, S.~Kajita, K.~Harada, K.~Kaneko, F.~Kanehiro,
  M.~Morisawa, and S.~Nakaoka, ``A pattern generator of humanoid robots walking
  on a rough terrain,'' in \emph{Robotics and Automation, 2007 IEEE
  International Conference on}.\hskip 1em plus 0.5em minus 0.4em\relax IEEE,
  2007, pp. 2181--2187.

\bibitem{zheng2010walking}
Y.~Zheng, M.~C. Lin, D.~Manocha, A.~H. Adiwahono, and C.-M. Chew, ``A walking
  pattern generator for biped robots on uneven terrains,'' in \emph{Intelligent
  Robots and Systems (IROS), 2010 IEEE/RSJ International Conference on}.\hskip
  1em plus 0.5em minus 0.4em\relax IEEE, 2010, pp. 4483--4488.

\bibitem{pang2000stability}
J.-S. Pang and J.~Trinkle, ``Stability characterizations of rigid body contact
  problems with coulomb friction,'' \emph{ZAMM-Journal of Applied Mathematics
  and Mechanics/Zeitschrift f{\"u}r Angewandte Mathematik und Mechanik},
  vol.~80, no.~10, pp. 643--663, 2000.

\bibitem{webpage}
\url{https://scaron.info/research/conf/icra-2015.html}, 2014, [Online].

\bibitem{pham2013kinodynamic}
Q.-C. Pham, S.~Caron, and Y.~Nakamura, ``Kinodynamic planning in the
  configuration space via admissible velocity propagation,'' in \emph{Robotics:
  Science and Systems}, 2013.

\bibitem{pham2014general}
\BIBentryALTinterwordspacing
Q.-C. Pham, ``A general, fast, and robust implementation of the time-optimal
  path parameterization algorithm,'' \emph{IEEE Transactions on Robotics (to
  appear)}, 2014. [Online]. Available: \url{http://arxiv.org/abs/1312.6533}
\BIBentrySTDinterwordspacing

\end{thebibliography}

\appendix

\section{Proof of Equation \ref{equaJ}}
\label{proof:equivwrench}

The rotation Jacobian satisfies $\bfomega = \Jrot \qd$, where $\bfomega$ is the
rotation velocity of the link. An interesting consequence of this property is
that for any point $C$ on the link and any vector $\bfu$, one has
\begin{equation}
    \label{Jacprop}
    \left(\tfrac{\partial\overrightarrow{OC}}{\partial \q}\right)^\top\bfu =
    \Jrot^\top\left(\overrightarrow{OC} \wedge \bfu\right).
\end{equation}
Next, the position of any point $C_i$ of the link in the absolute RF is related
to that of its origin $O$ by $C_i = O + \overrightarrow{OC_i}$. The
corresponding translation Jacobian is $ \bfJ(C_i) = \tfrac{\partial
C_i}{\partial \q} = \bfJ(O) + \tfrac{\partial \overrightarrow{OC_i}}{\partial
\q}$. Thus,
\begin{equation*}
    \textstyle
    \sum_i \bfJ(C_i)^\top \bff_i 
    = \bfJ(O)^\top \sum_i \bff_i+ \sum_i\left(\tfrac{\partial
    \overrightarrow{OC_i}}{\partial \q}\right)^\top \bff_i.
\end{equation*}
The first term of the expression equals the translation component of
$\bfJw^\top\bfW$. By applying \eqref{Jacprop}, we see that the second term
equals $\Jrot^\top (\overrightarrow{OC_i} \wedge \bff_i)$. Factoring
$\Jrot^\top$ out of the summation yields the rotation component of
$\bfJw^\top\bfW$. $\Box$

\subsection{Calculation of the Wrench Cone}
\label{bigproof}

\noindent The wrench is defined by \eqref{Wf-f}-\eqref{Wf-tau} as:
    \begin{equation*}
        \begin{array}{rcl}
        f^x & = & f_1^x + f_2^x + f_3^x + f_4^x \\
        f^y & = & f_1^y + f_2^y + f_3^y + f_4^y \\
        f^z & = & f_1^z + f_2^z + f_3^z + f_4^z \\
        \taux & = & Y (f^z_1 - f^z_2 - f^z_3 + f^z_4) \\
        \tauy & = & -X (f^z_1 + f^z_2 - f^z_3 - f^z_4) \\
        \tauz & = & X (f^y_1 + f^y_2 - f^y_3 - f^y_4) - Y (f^x_1 - f^x_2 -
        f^x_3 + f^x_4). 
    \end{array}
    \end{equation*} 
By unilaterality \eqref{fiz} we have $f^z > 0$, so we can define:
    \begin{equation*}
        \begin{array}{ccccccc}
            K_1 & := & f^x / \mu f^z &&
            C_1 & := & \taux / Y f^z \\
            K_2 & := & f^y / \mu f^z &&
            C_2 & := & \tauy / X f^z \\
            K_3 & := & \tauz / \mu (X+Y) f^z &&
            D_i & := & f_i^z / \sum_i f_i^z \\
            p_x & := & X / (X+Y) &&
            p_y & := & Y / (X+Y) \\
            \alpha_i^x & := & f_i^x / \mu f_i^z &&
            \alpha_i^y & := & f_i^y / \mu f_i^z \\
        \end{array}
    \end{equation*}
and normalize the system by dividing each row accordingly. From the
non-slippage constraint \eqref{fixy}, we have $\alpha_i^x, \alpha_i^y$
and $D_i \in [-1, 1]$. Then, introduce the new variables:
    \begin{equation*}
        \begin{array}{ccccccc}
            \gamma_x & = & \alpha_1^x D_1 + \alpha_4^x D_4  & &
        \gamma_x' & = & \alpha_2^x D_2 + \alpha_3^x D_3 \\
            \gamma_y & = & \alpha_1^y D_1 + \alpha_2^y D_2  & &
        \gamma_y' & = & \alpha_3^y D_3 + \alpha_4^y D_4 \\
        \end{array}
    \end{equation*}
We can reduce the complete system in two ways. First, using the fact that the
relation $\bfM$ from $\bfalpha$ to $\bfgamma$ is a linear surjection from $[-1,
1]^8$ to $\bfM [-1, 1]^8 = \{|\gamma_i^{x|y}| \leq D_j + D_k\}$ (the
computation of antecedents being straightforward). Since there is no other
constraint on the $\alpha_i$'s than their domain and relation to $\gamma_i$'s,
one can obtain an equivalent system by replacing $\bfalpha \in [-1, 1]^8$ by
$\bfgamma \in \bfM [-1, 1]^8$. Then, the three equations in $D_i$'s are:
\vspace{-.3em}
    \begin{equation*}
        \begin{array}{rcl}
        1 & = & D_1 + D_2 + D_3 + D_4, \\
        C_1 & = & D_1 - D_2 - D_3 + D_4, \\
        C_2 & = & -D_1 - D_2 + D_3 + D_4,
        \end{array}
    \end{equation*}
By linear combination, we can use them to rewrite $\bfM [-1, 1]^8$ as: $2
|\gamma_x| \leq 1 + C_1$, $2 |\gamma_x'| \leq 1 - C_1$, $2 |\gamma_y| \leq
1 - C_2$, $2 |\gamma_y'| \leq 1 + C_2$. Finally, using the same equations, one
can express all $D_i$'s as functions of, \eg $D_4$.  The inequalities $\bfD \in
[0, 1]^4$ become:
\vspace{-.3em}
    \begin{equation*}
        \begin{array}{ccccc}
            -1 + C_1 & \leq & 2 D_4 & \leq & 1 + C_1 \\
            C_1 + C_2 & \leq & 2 D_4 & \leq & 2 + C_1 + C_2  \\
            -1 + C_2 & \leq & 2 D_4 & \leq & 1 + C_2 \\
            0 & \leq & 2 D_4 & \leq & 2 \\
        \end{array}
    \end{equation*}
This system has solutions if and only if all lower bounds are smaller than all
upper bounds. Matching all pairs of lower and upper bounds one by one (we will
show an example of this technique below in a more complex situation), one can
check that all these inequalities boil down to $C_1 \in [-1, 1]$ and $C_2 \in
[-1, 1]$. The complete system is now:
\vspace{-.3em}
    \begin{equation*}
        \begin{array}{rclcrcl}
        K_1 & = & \gamma_x + \gamma_x' & &
        K_2 & = & \gamma_y + \gamma_y' \\
        K_3 & = & & \makebox[0pt]{$\quad \quad \quad p_x (\gamma_y
            - \gamma_y') - p_y (\gamma_x - \gamma_x')$} & &  \\
        2 |\gamma_x| & \leq & 1 + C_1  & &
        2 |\gamma_x'| & \leq & 1 - C_1 \\
        2 |\gamma_y| & \leq & 1 - C_2 & &
        2 |\gamma_y'| & \leq & 1 + C_2 \\
        \end{array}
    \end{equation*}
And $(C_1, C_2) \in [-1, 1]^2$. One can use the first three equations to
eliminate the redundant variables $\gamma_x'$, $\gamma_y$ and $\gamma_y'$,
expressing them as functions of $\gamma_x$ in the inequality constraints. After
simplification, the resulting system is:
\vspace{-.3em}
    \begin{eqnarray}
        2 p_y \gamma_x & \leq & p_y (1 + C_1)                             \label{h1} \\
        2 p_y \gamma_x & \leq & p_y (1 - C_1) + 2 p_y K_1                 \label{h2} \\
        2 p_y \gamma_x & \leq & p_x (1 - C_2) - K_3 + p_y K_1 - p_x K_2   \label{h3} \\
        2 p_y \gamma_x & \leq & p_x (1 + C_2) - K_3 + p_y K_1 + p_x K_2   \label{h4} \\
        2 p_y \gamma_x & \geq & -p_y (1 + C_1)                            \label{h5} \\
        2 p_y \gamma_x & \geq & -p_y (1 - C_1) + 2 p_y K_1                \label{h6} \\
        2 p_y \gamma_x & \geq & -p_x (1 - C_2) - K_3 + p_y K_1 - p_x K_2  \label{h7} \\
        2 p_y \gamma_x & \geq & -p_x (1 + C_2) - K_3 + p_y K_1 + p_x K_2  \label{h8}
    \end{eqnarray}
And $(C_1, C_2) \in [-1, 1]^2$. There exist a solution $\gamma_x$ if and only
if all of its lower bounds are smaller than all of its upper bounds. Let us
match all pairs of lower bounds \eqref{h5}-\eqref{h8} and upper bounds
\eqref{h1}-\eqref{h4}. One can check that:
\vspace{-.3em}
    \begin{itemize}
        \item {\footnotesize \eqref{h5} $\leq$ \eqref{h1}} $\Leftrightarrow$ {\footnotesize $C_1 \geq -1$}
        \item {\footnotesize \eqref{h5} $\leq$ \eqref{h2}} $\Leftrightarrow$ {\footnotesize $K_1 \geq -1$}
        \item {\footnotesize \eqref{h5} $\leq$ \eqref{h3}} $\Leftrightarrow$ {\footnotesize $K_3 - p_y K_1 + p_x K_2 - p_y C_1 + p_x C_2 \leq 1$}
        \item {\footnotesize \eqref{h5} $\leq$ \eqref{h4}} $\Leftrightarrow$ {\footnotesize $K_3 - p_y K_1 - p_x K_2 - p_y C_1 - p_x C_2 \leq 1$}

        \item {\footnotesize \eqref{h6} $\leq$ \eqref{h1}} $\Leftrightarrow$ {\footnotesize $K_1 \leq 1$}
        \item {\footnotesize \eqref{h6} $\leq$ \eqref{h2}} $\Leftrightarrow$ {\footnotesize $C_1 \leq 1$}
        \item {\footnotesize \eqref{h6} $\leq$ \eqref{h3}} $\Leftrightarrow$ {\footnotesize $K_3 + p_y K_1 + p_x K_2 + p_y C_1 + p_x C_2 \leq 1$}
        \item {\footnotesize \eqref{h6} $\leq$ \eqref{h4}} $\Leftrightarrow$ {\footnotesize $K_3 + p_y K_1 - p_x K_2 + p_y C_1 - p_x C_2 \leq 1$}

        \item {\footnotesize \eqref{h7} $\leq$ \eqref{h1}} $\Leftrightarrow$ {\footnotesize $-K_3 + p_y K_1 - p_x K_2 - p_y C_1 + p_x C_2 \leq 1$}
        \item {\footnotesize \eqref{h7} $\leq$ \eqref{h2}} $\Leftrightarrow$ {\footnotesize $-K_3 - p_y K_1 - p_x K_2 + p_y C_1 + p_x C_2 \leq 1$}
        \item {\footnotesize \eqref{h7} $\leq$ \eqref{h3}} $\Leftrightarrow$ {\footnotesize $C_2 \leq 1$}
        \item {\footnotesize \eqref{h7} $\leq$ \eqref{h4}} $\Leftrightarrow$ {\footnotesize $K_2 \geq -1$}

        \item {\footnotesize \eqref{h8} $\leq$ \eqref{h1}} $\Leftrightarrow$ {\footnotesize $-K_3 + p_y K_1 + p_x K_2 - p_y C_1 - p_x C_2 \leq 1$}
        \item {\footnotesize \eqref{h8} $\leq$ \eqref{h2}} $\Leftrightarrow$ {\footnotesize $-K_3 - p_y K_1 + p_x K_2 + p_y C_1 - p_x C_2 \leq 1$}
        \item {\footnotesize \eqref{h8} $\leq$ \eqref{h3}} $\Leftrightarrow$ {\footnotesize $K_2 \leq 1$}
        \item {\footnotesize \eqref{h8} $\leq$ \eqref{h4}} $\Leftrightarrow$ {\footnotesize $C_2 \geq -1$}
    \end{itemize}
\vspace{-.1em} Consequently, the complete system becomes:
\vspace{-.3em}
    \begin{equation*}
        \begin{array}{rcl}
        K_3 & \leq & 1 - p_y K_1 - p_x K_2 - p_y C_1 - p_x C_2  \\
        K_3 & \leq & 1 - p_y K_1 + p_x K_2 - p_y C_1 + p_x C_2  \\
        K_3 & \leq & 1 + p_y K_1 - p_x K_2 + p_y C_1 - p_x C_2  \\
        K_3 & \leq & 1 + p_y K_1 + p_x K_2 + p_y C_1 + p_x C_2  \\
        K_3 & \geq & -1 + p_y K_1 + p_x K_2 - p_y C_1 - p_x C_2 \\
        K_3 & \geq & -1 + p_y K_1 - p_x K_2 - p_y C_1 + p_x C_2 \\
        K_3 & \geq & -1 - p_y K_1 + p_x K_2 + p_y C_1 - p_x C_2 \\
        K_3 & \geq & -1 - p_y K_1 - p_x K_2 + p_y C_1 + p_x C_2
        \end{array}
    \end{equation*}
And $(K_1, K_2, C_1, C_2) \in [-1, 1]^4$. In a more concise form, these last
eight inequalities can be written {\small $K_3 \geq -1 + p_y |K_1 - C_1| + p_x
|K_2 - C_2|$} and {\small $K_3 \leq +1 - p_y |K_1 + C_1| - p_x |K_2 + C_2|$}.
We conclude by de-normalizing all inequalities. $\Box$

\end{document}